\documentclass[10pt,twocolumn,letterpaper]{article}

\usepackage{cvpr}
\usepackage{times}
\usepackage{epsfig}
\usepackage{graphicx}
\usepackage{amsmath}
\usepackage{amssymb}
\usepackage{enumitem}
\usepackage{multirow}
\usepackage{booktabs}
\usepackage{flushend}

\graphicspath{{figures/}}

\usepackage[pagebackref=true,breaklinks=true,letterpaper=true,colorlinks,bookmarks=false]{hyperref}

\cvprfinalcopy 

\def\cvprPaperID{2927} 

\ifcvprfinal\fi
\begin{document}

\title{Fine-grained Video-Text Retrieval with Hierarchical Graph Reasoning}

\author{Shizhe Chen\textsuperscript{1}\thanks{This work was performed when Shizhe Chen was visiting University of Adelaide.}, Yida Zhao\textsuperscript{1}, Qin Jin\textsuperscript{1}\thanks{Qin Jin is the corresponding author.}, Qi Wu\textsuperscript{2}\\
	\textsuperscript{1}School of Information, Renmin University of China\\ \textsuperscript{2}Australian Centre for Robotic Vision, University of Adelaide\\
	{\tt\small \{cszhe1, zyiday, qjin\}@ruc.edu.cn, qi.wu01@adelaide.edu.au}
}

\maketitle

\begin{abstract}
Cross-modal retrieval between videos and texts has attracted growing attentions due to the rapid emergence of videos on the web. 
The current dominant approach for this problem is to learn a joint embedding space to measure cross-modal similarities. 
However, simple joint embeddings are insufficient to represent complicated visual and textual details, such as scenes, objects, actions and their compositions. 
To improve fine-grained video-text retrieval, we propose a Hierarchical Graph Reasoning (HGR) model, which decomposes video-text matching into global-to-local levels.
To be specific, the model disentangles texts into hierarchical semantic graph including three levels of events, actions, entities and relationships across levels.
Attention-based graph reasoning is utilized to generate hierarchical textual embeddings, which can guide the learning of diverse and hierarchical video representations.
The HGR model aggregates matchings from different video-text levels to capture both global and local details.
Experimental results on three video-text datasets demonstrate the advantages of our model.
Such hierarchical decomposition also enables better generalization across datasets and improves the ability to distinguish fine-grained semantic differences.
\end{abstract}

\section{Introduction}
The rapid emergence of videos on the Internet such as on YouTube and TikTok has brought great challenges to accurate retrieval of video contents.
Traditional retrieval methods \cite{chang2015semantic,dalton2013zero,habibian2014composite} are mainly based on keyword search, where keywords are pre-defined and assigned to videos automatically or manually.
However, since keywords are limited and unstructured, it is difficult to retrieve various fine-grained contents, for example, accurately retrieving a video with a subject ``white dog" chasing an object ``black cat" is nearly impossible in keyword-based video retrieval system.
To address the limitation of keyword-based approach, more and more researchers are paying attention to video retrieval using natural language texts that contain richer and more structured details than keywords, a.k.a, cross-modal video-text retrieval~\cite{dong2019dual,mithun2018learning,yu2018joint}.

\begin{figure}
	\includegraphics[width=\linewidth]{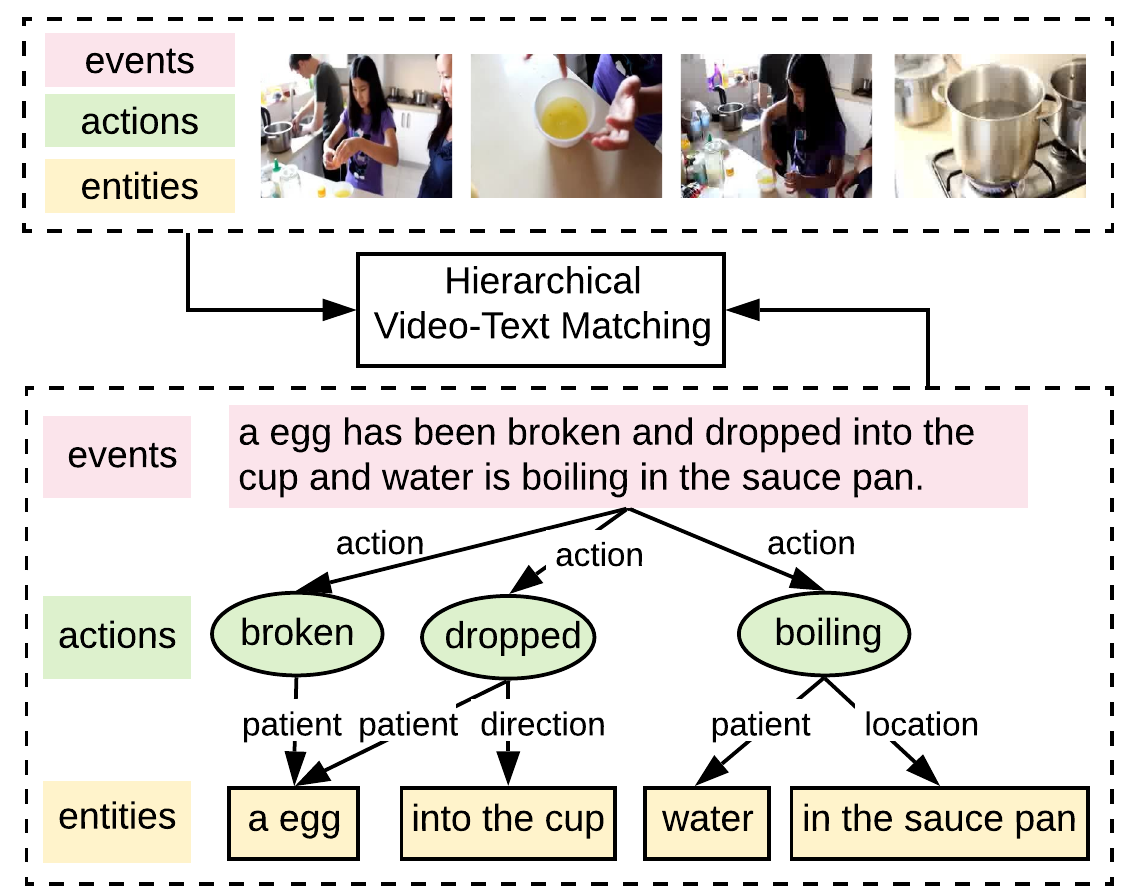}
	\caption{We factorize video-text matching into hierarchical levels including events, actions, and entities to form a global to local structure. On one hand, this enhances global matching with the help of detailed semantic components, on the other hand, it improves local matching with the help of global event structure.}
	\label{fig:intro}
	\vspace{-1em}
\end{figure}

The current dominant approach for cross-modal retrieval is to encode different modalities into a joint embedding space \cite{frome2013devise} to measure cross-modal similarities, which can be broadly classified into two categories.
The first type of works  \cite{dong2019dual,Liu19a,mithun2018learning,song2019polysemous} embeds videos and texts as global vectors that encode salient semantic meanings.
However, it can be hard for such a compact representation to capture fine-grained semantic details in texts and videos.
For example, understanding the video and text in Figure~\ref{fig:intro} involves complicated reasoning about different actions (break, drop, boil), entities (egg, into the cup \etc) as well as how all components compose to the event (`egg' is the patient of action `break' and `into the cup' is the direction).
To avoid losing those details, another type of methods \cite{song2019polysemous,yu2018joint} employs a sequence of frames and words to represent videos and texts respectively and aligns local components to compute overall similarities.
Although these approaches have achieved improved performance for image-text retrieval \cite{karpathy2015deep,lee2018stacked}, learning semantic alignments between videos and texts is more challenging since video-text pairs are more weakly supervised than image-text pairs.
Moreover, such sequential representations neglect topological structures in both videos and texts, so that they can not accurately capture relations between local components within an event and therefore may stuck in the local matching.

In this work, we propose a Hierarchical Graph Reasoning~(HGR)  model which takes the advantage of above global and local approaches and makes up their deficiencies.
As shown in Figure~\ref{fig:intro}, we decompose video-text matching into three hierarchical semantic levels, which are responsible to capture global events, local actions and entities respectively.
On the text side, the global event is represented by the whole sentence. Actions are denoted by verbs and entities refer to noun phrases.
Different levels are not independent and their interactions explain what semantic roles they play within the event. 
Therefore, we build a semantic role graph across levels in text and propose an attention-based graph reasoning method to capture such interaction.
Correspondingly, videos are encoded as hierarchical embeddings that relate to event, actions and entities to distinguish different aspects in videos.
We align cross-modal components at each semantic level via attention mechanisms to facilitate matching in weakly-supervised condition.
Matching scores from all three levels are aggregated together in order to enhance fine-grained semantic coverage.

We carry out extensive experiments on three video-text datasets including MSR-VTT \cite{xu2016msr}, TGIF \cite{li2016tgif} and VATEX \cite{Wang_2019_ICCV}.
The consistent improvements over previous global and local approaches demonstrate the effectiveness of the proposed Hierarchical Graph Reasoning model.
Our hierarchical decomposition mechanism also enables better generalization ability when directly applying pre-trained models to an unseen dataset Youtube2Text \cite{guadarrama2013youtube2text}.
To further evaluate our model on fine-grained retrieval, we propose a new binary selection task \cite{hodosh2016focused,hu2019evaluating} 
which requires the system to select the correct matching sentence for the given video from two similar candidate sentences with subtle difference.
Our model achieves better performance on recognizing various fine-grained semantic changes such as switching roles, replacing items etc. of ground-truth video descriptions.
What is more, due to the fusion of hierarchical matching, our HGR model can prefer the comprehensive video description than the ones that are correct but incomplete.

The contributions of this work are as follows:
\parskip=0.1em
\begin{itemize}[itemsep=0.1em,parsep=0em,topsep=0em,partopsep=0em]
	\item We propose a Hierarchical Graph Reasoning (HGR) model that decomposes video-text matching into global-to-local levels. It improves global matching with the help of detailed semantics and local matching with the help of global event structures for fine-grained video-text retrieval.
	\item The three disentangled levels in texts such as event, actions and entities are interacted with each other via attention-based graph reasoning and aligned with corresponding levels of videos. All levels are contributed to video-text matching for better semantic coverage.
	\item The HGR model achieves improved performance on different video-text datasets and better generalization ability on unseen dataset. A new binary selection task to demonstrate the ability to distinguish fine-grained semantic differences is proposed as well. 
\end{itemize}

\section{Related Works}

\noindent\textbf{Image-Text Matching.}
Most of previous works \cite{faghri2018vse++,frome2013devise,gu2018look,huang2018learning,kiros2014unifying} for image-text matching encode images and sentences as fix-dimensional vectors in a common latent space for similarity measure.
Frome \etal \cite{frome2013devise} firstly propose the joint embedding framework for images and words, and train the model with contrastive ranking loss.
Kiros \etal \cite{kiros2014unifying} extend the framework to match images and sentences with CNN to encode images and RNN for sentences.
Faghri \etal \cite{faghri2018vse++} improve training strategy with hard negative mining.
To enrich global representations, Huang \etal \cite{huang2018learning} utilize image embeddings to predict concepts and orders via image captioning, and Gu \etal \cite{gu2018look} further incorporate image and caption generation in multi-task framework.
However, it is hard to cover complicated semantics only using fixed-dimensional vectors.
Therefore, Karpathy \etal~\cite{karpathy2015deep} decompose image and sentences as multiple regions and words, and propose using maximum alignment to compute global matching similarity.
Lee \etal \cite{lee2018stacked} improve the alignment with stacked cross-attention.
Wu \etal \cite{wu2019unified} factorize sentence into objects, attributes, relations and sentences, however, they do not consider interactions between different levels and the decomposition might not be optimal for video descriptions that focus on actions and events. 

\begin{figure*}
	\includegraphics[width=\linewidth]{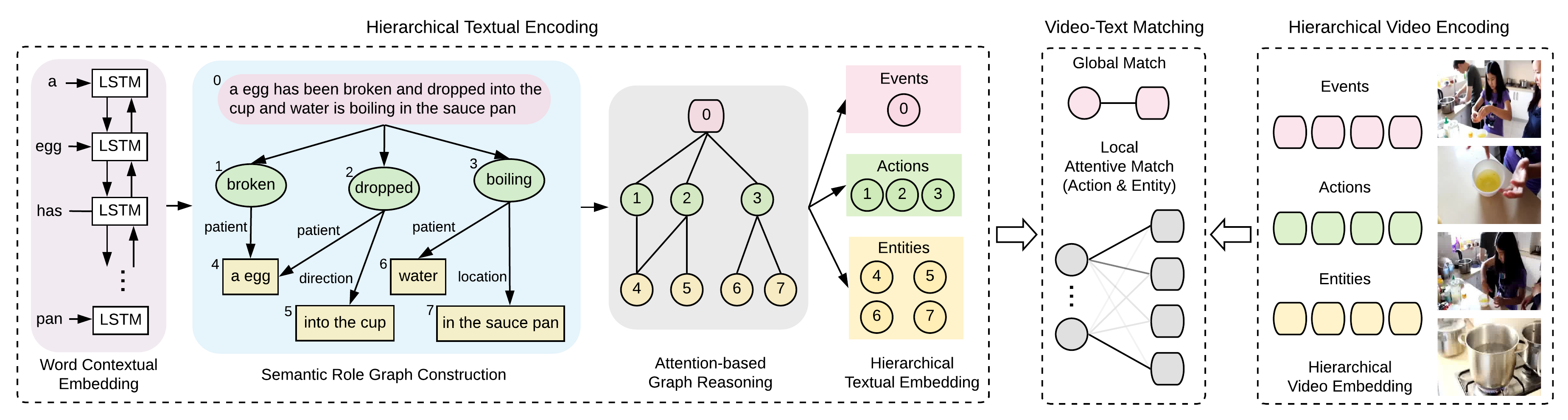}
	\caption{Overview of the proposed Hierarchical Graph Reasoning (HGR) model for cross-modal video-text retrieval.}
	\label{fig:method}
	\vspace{-1em}
\end{figure*}

\noindent\textbf{Video-Text Matching.}
Though sharing certain similarities with image-text matching, the video-text matching task is more challenging because videos are more complicated with multi-modalities and spatial-temporal evolution.
Mithun \etal \cite{mithun2018learning} employ multimodal cues from image, motion, audio modalities in video.
Liu \etal \cite{Liu19a} further utilize all modalities that can be extracted from videos such as speech contents and scene texts for video encoding.
In order to encode sequential videos and texts, Dong \etal \cite{dong2019dual} utilize three branches, \emph{i.e.} mean pooling, biGRU and CNN to encode them.
Yu \etal \cite{yu2018joint} propose a joint sequence fusion model for sequential interaction of videos and texts.
Song \etal \cite{song2019polysemous} employ multiple diverse representations for videos and texts for the polysemous problem.
The most similar work to ours is Wray \etal \cite{Wray_2019_ICCV}, which disentangles action phrases into different part-of-speech such as verbs and nouns for fine-grained action retrieval.
However, sentences are more complicated than action phrases. Therefore, in this work we decompose a sentence as a hierarchical semantic graph and integrate video-text matching at different levels.

\noindent\textbf{Graph-based Reasoning.}
The graph convolutional network (GCN) \cite{kipf2016semi} is firstly proposed for graph data recognition. For each node, it employs convolution on its neighbourhoods as outputs.
Graph attention networks \cite{velivckovic2017graph} are further introduced to dynamically attend over neighborhoods' features
In order to model graphs with different edge types, relational GCN is proposed in \cite{schlichtkrull2018modeling} that learns specific contextual transformation for each relation type.
The graph-based reasoning has great applications in computer vision tasks such as action recognition \cite{sun2018actor,wang2018videos}, scene graph generation \cite{yang2018graph}, referring expression grounding \cite{li2019visual,wang2019neighbourhood}, and visual question answering \cite{hu2019language,li2019relation} etc.
Most of them \cite{hu2019language,li2019visual,li2019relation,wang2019neighbourhood,yang2018graph} apply graph reasoning on image regions to learn relationships among them.
In this work, we focus on reasoning over hierarchical graph structures on video descriptions for fine-grained video-text matching.

\section{Hierarchical Graph Reasoning Model}
Figure~\ref{fig:method} illustrates the overview of the proposed HGR model which consists of three blocks: 
1) hierarchical textual encoding (Section~\ref{sec:text_encoder}) that constructs semantic role graphs from texts and applies graph reasoning to obtain hierarchical text representations, 
2) hierarchical video encoding (Section~\ref{sec:video_encoder}) that maps videos into corresponding multi-level representations, 
and 3) video-text matching (Section~\ref{sec:video_text_matching}) which aggregates global and local matching at different levels to compute overall cross-modal similarities. 

\subsection{Hierarchical Textual Encoding}
\label{sec:text_encoder}
Video descriptions naturally contain hierarchical structures.
The overall sentence describes the global event in the video which is composed of multiple actions in temporal dimensions, and each action is composed of different entities as its arguments such as agent and patient of the action.
Such global-to-local structure is beneficial to accurately and comprehensively understand the semantic meanings of video descriptions.
Therefore, in this section, we introduce how to obtain hierarchical textual representations from a video description in a global-to-local topology.

\noindent\textbf{Semantic Role Graph Structure.}
Given a video description $C$ that consists of $N$ words $\{c_{1}, \cdots, c_{N} \}$, we consider $C$ as a global event node in the hierarchical graph.
Then we employ an off-the-shelf semantic role parsing toolkit \cite{shi2019simple} to obtain verbs, noun phrases in $C$ as well as the semantic role of each noun phrase to the corresponding verb (details of semantic roles are given in the supplementary).
The verbs are considered as action nodes and connected to event node with direct edges, so that temporal relations of different actions can be implicitly learned from event node in following graph reasoning.
The noun phrases are entity nodes that are connected with different action nodes.
The edge type between action and entity nodes is decided by the semantic role of the entity in reference to the action.
If an entity node servers multiple semantic roles to different action nodes, we duplicate the entity node for each semantic role.
Such semantic role relations are important to understand the event structure, for example, ``a dog chasing a cat'' is apparently different from ``a cat chasing a dog'' which only switches the semantic roles of the two entities.
In the left side of Figure~\ref{fig:method}, we present an example of the constructed graph.

\noindent\textbf{Initial Graph Node Representation.}
We embed semantic meaning of each node into a dense vector as initialization.
For the global event node, we aim to summarize salient event described in the sentence. Therefore, we first utilize an bidirectional LSTM (Bi-LSTM) \cite{hochreiter1997long} to generate a sequence of contextual-aware word embeddings $\{w_1, \cdots, w_{N} \}$ as follows:

\begin{eqnarray}
	\overrightarrow{w}_i &=& \overrightarrow{\mathrm{LSTM}} (W_c c_i, \overrightarrow{w}_{i-1};\overrightarrow{\theta} ) \\
	\overleftarrow{w}_i &=& \overleftarrow{\mathrm{LSTM}} (W_c c_i, \overleftarrow{w}_{i+1}; \overleftarrow{\theta}) \\
	w_i &=& (\overrightarrow{w}_i +  \overleftarrow{w}_i) / 2
\end{eqnarray}
where $W_c$ is word embedding matrix, $\overrightarrow{\theta}$ and $ \overleftarrow{\theta}$ are parameters in the two LSTMs.
Then we average word embeddings via an attention mechanism that focuses on important words in the sentence as the global event embedding $g_e$:
\begin{eqnarray}
\label{eqn:event_attn}
g_e & = & \sum_{i=1}^{N} \alpha_{e, i} w_i \\
\alpha_{e, i} & = & \frac{\mathrm{exp}(W_{e} w_i )}{\sum_{j=1}^{N} \mathrm{exp}(W_{e} w_j )}
\end{eqnarray}
where $W_{e}$ is the parameter to be learned.
For action and entity nodes, though different LSTMs can be employed to only encode their constitutive words independently, since semantic role parsing might separate words with mistakes, contextual word representations can be beneficial to resolve such negative influences.
Therefore, we reuse the above Bi-LSTM word embeddings $w_i$ and apply max pooling over words in each node as action node representations $g_{a} = \{g_{a, 1}, \cdots, g_{a, N_a} \}$ and entity node representations $g_{o} = \{g_{o, 1}, \cdots, g_{o, N_o} \}$, where $N_a$ and $N_o$ are numbers of action and entity nodes respectively.

\noindent\textbf{Attention-based Graph Reasoning.}
The connections across different levels in the constructed graph not only explain how local nodes compose the global event, but also are able to reduce ambiguity for each node.
For example, the entity ``egg" in Figure~\ref{fig:method} can have diverse appearances without context, but the context from action ``break'' constrains its semantics, so that it should have high similarity with visual appearance of a ``broken egg'' rather than a ``round egg''.
Therefore, we propose to reason over interactions in the graph to obtain hierarchical textual representations.

Since edges in our graph are of different semantic roles, a straightforward approach to model interactions in graph is to utilize relational GCN \cite{schlichtkrull2018modeling}, which requires to learn separate transformation weight matrix for each semantic role.
However, it can lead to rapid growth of parameters, which makes it inefficient to learn from limited amount of video-text data and prone to over-fitting on rare semantic roles.

To address this problem, we propose to factorize multi-relational weights in GCN into two parts: a common transformation matrix $W_t \in \mathbb{R}^{D \times D}$ that is shared for all relationship types and a role embedding matrix $W_r \in \mathbb{R}^{D \times K}$ that is specific for different semantic roles, where $D$ is the dimension of node representation and $K$ is the number of semantic roles.
For inputs to the first GCN layer, we multiply initialized node embeddings $g_i \in \{g_e, g_a, g_o\}$ with their corresponding semantic roles as:
\begin{equation}
\label{eqn:role_aware}
  g_{i}^{0} = g_i \odot W_r r_{ij}
\end{equation}
where $r_{ij}$ is an one-hot vector denoting the type of semantic role between node $i$ and $j$.
Suppose $g_i^{l}$ is the output representation of node $i$ at $l$-th GCN layer, we employ a graph attention network to select relevant contexts from neighbor nodes to enhance the representation for each node:
\begin{eqnarray}
\tilde{\beta}_{ij} & = & (W_a^{q} g_i^{l})^{T} (W_a^{k} g_j^{l}) / \sqrt{D} \\
\beta_{ij} & = & \frac{\mathrm{exp} (\tilde{\beta}_{ij})}{\sum_{j \in \mathcal{N}_{i}} \mathrm{exp} (\tilde{\beta}_{ij}) }
\end{eqnarray}
where $\mathcal{N}_i$ is neighborhood nodes of node $i$, $W_a^{k}$ and $W_a^{q}$ are parameters to compute graph attention.
Then the shared $W_r$ is utilized to transform contexts from attended nodes to node $i$ with residual connection:
\begin{equation}
\label{eqn:gan}
	g_i^{l+1} = g_i^{l} + W_t^{l+1} \sum_{j \in \mathcal{N}_{i}} (\beta_{ij} g_j^{l})
\end{equation}
Putting together Eq~(\ref{eqn:role_aware}) and Eq~(\ref{eqn:gan}), we can see that the transformation from nodes in lower layer is specific for different semantic role edges.
Take the first GCN layer as an example, the computation is as follows:
\begin{equation}
	g_i^1 = g_i^0 + \sum_{j \in \mathcal{N}_i} (\beta_{ij} (W_t^{1} \odot W_r r_{ij})  g_j)
\end{equation}
where $\odot$ is element-wise multiplication with broadcasting, $W_t^{1} \odot W_r r_{ij}$ is the edge specific transformation at layer 1.
In this way, we significantly reduce the size of parameters from $L \times K \times D \times D$ to $L \times D \times D + K \times D$ where $L$ is the number of layers of GCN, but still maintain role-awareness when reasoning over graph.
The outputs from the $L$-th GCN layer are our final hierarchical textual representations, which are denoted as $c_e$ for global event node, $c_a$ for action nodes and $c_o$ for entity nodes.

\subsection{Hierarchical Video Encoding}
\label{sec:video_encoder}
Videos also contain multiple aspects such as objects, actions and events. 
However, it is challenging to directly parse video into hierarchical structures as in texts which requires temporal segmentation, object detection, tracking and so on.
We thus instead build three independent video embeddings to focus on different level of aspects in the video.

Given video $V$ as a sequence of frames $\{f_1, \cdots, f_M \}$, we utilize different transformation weights $W^v_e, W^v_a$ and $W^v_o$ to encode videos into three level of embeddings:
\begin{equation}
	v_{x, i} = W^v_x f_i, \quad x \in \{e, a, o\}
\end{equation}
For the global event level, we employ the attention mechanism similar to Eq~(\ref{eqn:event_attn}) to obtain one global vector to represent the salient event in the video as $v_e$.
And for the action and entity level, the video representations are a sequence of frame-wise features $v_a = \{v_{a, 1}, \cdots, v_{a, M} \}$ and $v_o = \{v_{o, 1}, \cdots, v_{o, M} \}$ respectively. These features will be sent to the following matching module to match with their corresponding textual features at different levels, which guarantees different transformation weights can be learned to focus on different level video information via an end-to-end learning fashion.

\begin{table*}
	\centering
	\caption{Cross-modal retrieval comparison with state-of-the-art methods on MSR-VTT testing set.}
	\label{tab:msrvtt_sota}
	\begin{tabular}{l|ccccc|ccccc|c} \toprule
		\multirow{2}{*}{Model} & \multicolumn{5}{c|}{Text-to-Video Retrieval} & \multicolumn{5}{c|}{Video-to-Text Retrieval} & \multirow{2}{*}{rsum} \\
		& R@1 & R@5 & R@10 & MedR & MnR & R@1 & R@5 & R@10 & MedR & MnR &  \\ \midrule
		VSE \cite{kiros2014unifying} & 5.0 & 16.4 & 24.6 & 47 & 215.1 & 7.7 & 20.3 & 31.2 & 28 & 185.8 & 105.2 \\
		VSE++ \cite{faghri2018vse++} & 5.7 & 17.1 & 24.8 & 65 & 300.8 & 10.2 & 25.4 & 35.1 & 25 & 228.1 &  118.3 \\
		Mithum \etal \cite{mithun2018learning} & 5.8 & 17.6 & 25.2 & 61 & 296.6 & 10.5 & 26.7 & 35.9 & 25 & 266.6 & 121.7 \\
		W2VV \cite{dong2018predicting} & 6.1 & 18.7 & 27.5 & 45 & - & 11.8 & 28.9 & 39.1 & 21 & - & 132.1 \\
		Dual Encoding \cite{dong2019dual} & 7.7 & 22.0 & 31.8 & 32 & - & 13.0 & 30.8 & 43.3 & 15 & - & 148.6 \\ \midrule
		Our HGR & \textbf{9.2} & \textbf{26.2} & \textbf{36.5} & \textbf{24} & \textbf{164.0} & \textbf{15.0} & \textbf{36.7} & \textbf{48.8} & \textbf{11} & \textbf{90.4} & \textbf{172.4} \\ \bottomrule
	\end{tabular}
\end{table*}

\subsection{Video-Text Matching}
\label{sec:video_text_matching}
In order to cover both local and global semantics to match videos and texts, we aggregate results from the three hierarchical levels for the overall cross-modal similarity.

\noindent\textbf{Global Matching.}
At the global event level, the video and text are encoded into global vectors that capture salient event semantics with attention mechanism.
Therefore, we simply utilize cosine similarity $cos(v, c) \equiv \frac{v^T c}{||v||  ||c||}$ to measure the cross-modal similarity for global video and text contents.
The global matching score is $s_e = cos(v_e, c_e)$.

\noindent\textbf{Local Attentive Matching.}
At the action and entity level, there are multiple local components in the video and text.
Therefore, an alignment between cross-modal local components is supposed to be learned to compute overall matching score.
For each $c_{x, i} \in c_x$ where~$x \in \{a, o\}$, we first compute local similarities between each pair of cross-modal local components $s_{ij}^{x} = cos(v_{x, j}, c_{x, i})$.
Such local similarities implicitly reflect the alignment between local texts and videos such as how strong a text node is relevant to a video frame, but they lack proper normalization.
Therefore, we normalize $s_{ij}^{x}$ inspired by stacked attention \cite{lee2018stacked} as follows:
\begin{equation}
\varphi_{ij}^{x} = \mathrm{softmax}(\lambda ([s_{ij}^{x}]_{+} / \sqrt{\sum_j [s_{ij}^{x}]_{+}^{2}}))
\end{equation}
where $[\cdot]_{+} \equiv \max(\cdot, 0)$.
The $\varphi_{ij}^{x}$  is then utilized as attention weights over video frames for each local textual node $i$, which dynamically aligns $c_{x, i}$ to video frames.
We then compute the similarity between $c_{x, i}$ and $v_x$ as weighted average of local similarities $s_{x, i} = \sum_j \varphi_{ij}^{x} s_{ij}^{x}$.
The final matching similarity summarizes all local component similarities of text $s_{x} = \sum_{i} s_{x, i}$.
The local attentive matching does not require any local text-video groundings, and can be learned from the weakly supervised global video-text pairs.

\noindent\textbf{Training and Inference.}
We take the average of cross-modal similarities at all levels as final video-text similarity:
\begin{equation}
	s(v, c) = (s_e + s_a + s_o) / 3
\end{equation}
The contrastive ranking loss is employed as training objective. For each positive pair $(v^{+}, c^{+})$, we find its hardest negatives in a mini-batch $(v^{+}, c^{-})$ and $(v^{-}, c^{+})$, and push their distances from the positive pair $(v^{+}, c^{+})$ further away than a pre-defined margin $\Delta$ as follows:
\begin{equation}
\begin{split}
  L(v^{+}, c^{+}) = [\Delta + s(v^{+}, c^{-}) - s(v^{+}, c^{+})]_{+} \\
  +  [\Delta + s(v^{-}, c^{+}) - s(v^{+}, c^{+})]_{+}
\end{split}
\end{equation}

\section{Experiments}
To demonstrate the effectiveness of our HGR model, we compare it with state-of-the-art (SOTA) methods on three video-text datasets for text-to-video retrieval and video-to-text retrieval.
Extensive ablation studies are conducted to investigate each component of our model.
We also propose a binary selection task to evaluate fine-grained discrimination ability of different models for cross-modal retrieval.

\begin{table*}
	\centering
	\caption{Generalization on unseen Youtute2Text testing set using different pre-trained models on MSR-VTT dataset.}
	\label{tab:msvd_generalize}
	\begin{tabular}{l|ccccc|ccccc|c} \toprule
		\multirow{2}{*}{Model} & \multicolumn{5}{c|}{Text-to-Video Retrieval} & \multicolumn{5}{c|}{Video-to-Text Retrieval} & \multirow{2}{*}{rsum} \\ 
		& R@1 & R@5 & R@10 & MedR & MnR & R@1 & R@5 & R@10 & MedR & MnR &  \\ \midrule
		VSE \cite{kiros2014unifying} & 11.0  & 28.6 & 39.9 & 18  & 48.7 & 15.4 & 31.0 & 42.4 & 19 & 128.0 & 168.3 \\
		VSE++ \cite{faghri2018vse++} & 13.8 & 34.6 & 46.1 & 13 & 48.4 & 20.8 & 37.6 & 47.8 & 12 & 108.3 & 200.6 \\
		Dual Encoding \cite{dong2019dual} & 12.7 & 32.0 & 43.8 & 15 & 52.7 & 18.7 & 37.2 & 45.7 & 15 & 142.6 & 190.0 \\
		Our HGR & \textbf{16.4} & \textbf{38.3} & \textbf{49.8} & \textbf{11} & \textbf{49.2} & \textbf{23.0} & \textbf{42.2} & \textbf{53.4} & \textbf{8} & \textbf{77.8} & \textbf{223.2} \\ \bottomrule
	\end{tabular}
\end{table*}

\begin{table}
	\centering
	\small
	\caption{Text-to-video retrieval comparison with state-of-the-art methods on TGIF and VATEX testing set.}
	\label{tab:tgif_vatex_sota}
	\begin{tabular}{c|l|cccc} \toprule
		Dataset & Model & R@1 & R@5 & R@10 &MedR \\ \midrule
		\multirow{6}{*}{TGIF} & DeViSE \cite{frome2013devise} & 0.8 & 3.5 & 6.0 & 379 \\
		& VSE++ \cite{faghri2018vse++} & 0.4 & 1.6 & 3.6 & 692 \\
		& Order \cite{vendrov2015order} & 0.5 & 2.1 & 3.8 & 500 \\
		& Corr-AE \cite{feng2014cross} & 0.9 & 3.4 & 5.6 & 365 \\
		& PVSE \cite{song2019polysemous} & 2.3 & 7.5 & 11.9 & 162 \\
		& HGR & \textbf{4.5} & \textbf{12.4} & \textbf{17.8} & \textbf{160} \\ \midrule
		\multirow{4}{*}{VATEX} & VSE \cite{kiros2014unifying} & 28.0 & 64.2 & 76.9 & 3 \\
		& VSE++ \cite{faghri2018vse++} & 33.7 & 70.1 & 81.0 & \textbf{2} \\
		& \footnotesize{Dual Encoding} \cite{dong2019dual}& 31.1 & 67.4 & 78.9 & 3 \\
		& HGR & \textbf{35.1} & \textbf{73.5} & \textbf{83.5} & \textbf{2} \\ \bottomrule
	\end{tabular}
\end{table}

\subsection{Experimental Settings}
\noindent\textbf{Datasets.}
We carry out experiments on MSR-VTT \cite{xu2016msr}, TGIF \cite{li2016tgif} and recent VATEX \cite{wang2019vatex} video-text datasets.
The MSR-VTT dataset contains 10,000 videos with 20 text descriptions for each video.
We follow the standard split with 6,573 videos for training, 497 for validation and 2,990 for testing.
The TGIF dataset contains gif format videos, where there are 79,451 videos for training, 10,651 for validation and 11,310 for testing in the official split \cite{li2016tgif}.
Each video is annotated with 1 to 3 text descriptions.
The VATEX dataset includes 25,991 videos for training, 3,000 for validation and 6,000 for testing.
Since the annotations on testing set are private, we randomly split the validation set into two equal parts with 1,500 videos as validation set and other 1,500 videos as our testing set.
There are 10 sentences in English and Chinese languages to describe each video. In this work, we only utilize the English annotations.

\noindent\textbf{Evaluation Metrics.}
We measure the retrieval performance with common metrics in information retrieval, including Recall at K (R@K), Median Rank (MedR) and Mean Rank (MnR).
R@K is the fraction of queries that correctly retrieve desired items in the top K of ranking list. We utilize K = 1, 5, 10 following the tradition.
The MedR and MnR measures the median and average rank of correct items in the retrieved ranking list respectively, where lower score indicates a better model.
We also take the sum of all R@K as rsum to reflect the overall retrieval performance.

\noindent\textbf{Implementation Details.}
For the video encoding, we use Resnet152 pretrained on Imagenet \cite{he2016deep} to extract frame-wise features for MSR-VTT and TGIF.
We utilize the officially provided I3D  \cite{i3d} video feature for VATEX dataset.
For the text encoding, we set the word embedding size as 300 and initialize with pretrained Glove embeddings \cite{pennington2014glove}. 
We use two layers of attentional graph convolutions. The dimension of joint embedding space for each level is 1024.
We set $\lambda = 4$ in local attentive matching.
For training, we set the margin $\Delta = 0.2$, and train the model for 50 epochs with mini-batch size of 128.
The epoch with the best rsum on validation set is selected for inference.

\begin{table*}
	\centering
	\caption{Ablation studies on MSR-VTT dataset to investigate contributions of different components of our HGR model.}
	\label{tab:ablation_msrvtt}
	\begin{tabular}{l|c|ccccc|ccccc|c} \toprule
		& \multirow{2}{*}{Model} & \multicolumn{5}{c|}{Text-to-Video Retrieval} & \multicolumn{5}{c|}{Video-to-Text Retrieval} & \multirow{2}{*}{rsum} \\
		& & R@1 & R@5 & R@10 & MedR & MnR & R@1 & R@5 & R@10 & MedR & MnR &  \\ \midrule
		1 & w/o graph attention & 8.9 & 25.3 & 35.6 & 25 & 173.5 & 14.5 & 35.7 & 47.1 & 12 & 96.5 & 167.1 \\
		2 & w/o role awareness & 9.1 & 25.7 & 36.3 & 24 & 171.3 & 14.2 & 34.7 & 46.8 & 12 & 98.0 & 166.8 \\
		3 & w/o hierarchical video & 8.8 & 25.5 & 36.2 & 24 & 170.2 & \textbf{15.2} & 35.1 & 47.2 & 12 & 108.9 & 167.9 \\ \midrule
		4 & full HGR model & \textbf{9.2} & \textbf{26.2} & \textbf{36.5} & \textbf{24} & \textbf{164.0} & 15.0 & \textbf{36.7} & \textbf{48.8} & \textbf{11} & \textbf{90.4} & \textbf{172.4} \\\bottomrule
	\end{tabular}
\end{table*}

\subsection{Comparison with State of The Arts}
Table~\ref{tab:msrvtt_sota} compares the proposed HGR model with SOTA methods on the MSR-VTT testing set. 
For fair comparison, all the models utilize the same video features.
Our model achieves the best performance across different evaluation metrics on the MSR-VTT dataset.
It outperforms the state-of-the-art Dual Encoding~\cite{dong2019dual} method even with half less parameters and computations, which obtains 19.5\% and 15.4\% relative gains on R@1 metric for text-to-video and video-to-text retrieval respectively.
The overall retrieval quality reflected by the rsum metric is also boosted by a large margin (+23.8). 
We believe the major gain comes from our global-to-local matching and attention-based graph reasoning to learn hierarchical textual representations.
Though Dual Encoding enhances global video and sentence features via ensembling different networks such as mean pooling, RNNs and CNNs, it may still focus on the global event level and thus not as efficient as ours to capture fine-grained semantic details in text for cross-modal video-text retrieval.

To demonstrate the robustness of our approach on different datasets and features, we further provide quantitative results on TGIF and VATEX datasets in Table~\ref{tab:tgif_vatex_sota}.
The models employ Resnet152 image features on the TGIF dataset and I3D video features on the VATEX dataset.
We can see that our HGR model achieves consistent improvements across different datasets and features compared to SOTA models, which demonstrates that it is beneficial to improve the cross-modal retrieval accuracy by decomposing videos and texts into global-to-local hierarchical graph structures.

\subsection{Generalization on Unseen Dataset}
Current video-text retrieval methods are mainly evaluated on the same dataset. However, it is important for the model to generalize to out-of-domain data.
Therefore, we further conduct generalization evaluations: we first pretrain a model on one dataset and then measure its performance on another dataset that is unseen in the training.
Specifically, we utilize the MSR-VTT dataset for training and test models on the Youtube2Text testing split \cite{guadarrama2013youtube2text}, which contains 670 videos and 41.5 descriptions per video on average.

Table~\ref{tab:msvd_generalize} presents retrieval results on the Youtube2Text dataset.
The hard negative training strategy proposed in VSE++~\cite{faghri2018vse++} enables the model to learn visual-semantic matching more effectively, which also improves model's generalization ability on unseen data.
The Dual Encoding model though achieves better retrieval performance on the MSR-VTT dataset as show in Table~\ref{tab:msrvtt_sota}, it does not generalize well on a new dataset compared with VSE++ with overall 10.6 points decrease on rsum metric.
Our HGR model instead not only outperforms previous approaches on in-domain evaluation, but also achieves significantly better retrieval performance on out-of-domain dataset.
This property proves that improvements of our model does not result from using more complicated networks that might over-fit datasets.
Since we decompose texts into structures of events, actions and entities from global to local and match them with hierarchical video embeddings, our model is capable of learning better alignments of local components as well as global event structures, which improves the generalization ability on new compositions.

\subsection{Ablation Studies }
\begin{figure}
	\includegraphics[width=1\linewidth]{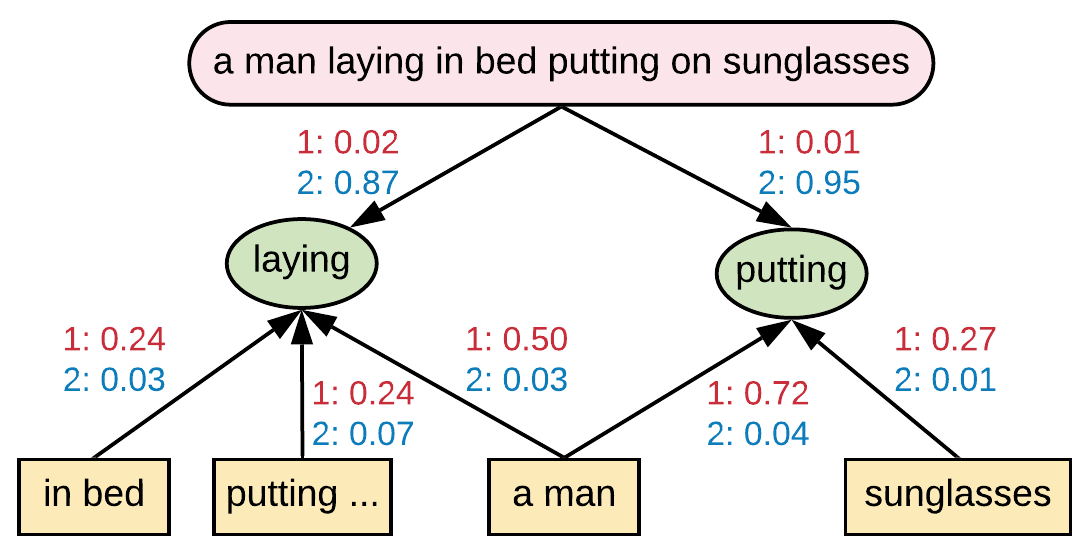}
	\caption{The attention distributions of action nodes at different graph reasoning layers to gather contexts from other nodes. The number in red after 1 denotes attention score in the first attention layer, while the number in blue after 2 denotes attention score in the second attention layer.}
	\label{fig:graph_attn_example}
	\vspace{-1em}
\end{figure}

In order to investigate contributions of different components in our proposed model, we carry out ablation studies on the MSR-VTT dataset in Table~\ref{tab:ablation_msrvtt}.
The Row 1 in Table~\ref{tab:ablation_msrvtt} replaces graph attention mechanism in graph reasoning and simply utilizes average pooling over neighbor nodes, which reduces the retrieval performance with 0.9 and 1.7 on R@10 metric than the full model in Row 4 for text-to-video and video-to-text retrieval respectively.
The role awareness in Eq~(\ref{eqn:role_aware}) is also beneficial in graph reasoning comparing Row 2 and Row 4, which enables the model to understand how different components relate with each other within an event.
In Figure~\ref{fig:graph_attn_example}, we present a learned pattern on how action nodes interacting with neighbor nodes in graph reasoning at different layers, which is strongly relevant to semantic roles.
At the first attention layer, the action node such as ``laying'', ``putting'' focuses more on its main arguments such as agent ``man''.
Then at the second layer, action nodes begin to reason over their temporal relations and thus pay more attention to temporal arguments as well as implicit contexts from global event node.

We also show that representing videos as hierarchical embeddings is important to capture different aspects in the video, which improves overall rsum performance from 167.9 in Row 3 to 172.4 in row 4.
Since our video-text similarities are aggregated from different levels, in Table~\ref{tab:level_results_msrvtt} we break down the performance at each level for video-text retrieval.
We can see that the global event level performs the best alone on rsum metric since local levels might not contain overall event structures on itself.
But different levels are complementary with each other and their combinations significantly improves the retrieval performance.

\begin{table}
	\centering
	\caption{Break down of retrieval performance at different levels on MSR-VTT testing set.}
	\label{tab:level_results_msrvtt}
	\begin{tabular}{c|ccc|ccc} \toprule
		\multirow{2}{*}{} & \multicolumn{3}{c|}{Text-to-Video} & \multicolumn{3}{c}{Video-to-Text} \\
		& rsum & MedR & MnR & rsum & MedR & MnR \\ \midrule
		event & 57.6 & 43 & 267.8 & 77.8 & 20.5 & 258.0 \\
		action & 50.4 & 77 & 441.6 & 80.7 & 22 & 241.4 \\
		entity & 44.7 & 62 & 251.3 & 58.4 & 37 & 230.0 \\ \midrule
		fusion & \textbf{71.9} & \textbf{24} & \textbf{164.0} & \textbf{100.6} & \textbf{11} & \textbf{90.4} \\ \bottomrule
	\end{tabular}
\end{table}

\begin{table*}
	\centering
	\caption{Performance of different models on fine-grained binary selection task.}
	\label{tab:binary_task}
	\begin{tabular}{l|ccccc|c} \toprule
		Model & switch roles & replace actions & replace persons & replace scenes & incomplete events & average \\ \midrule
		\# of triplets & 616 & 646 & 670 & 539 & 646 & 623.4 \\ \midrule
		VSE++ \cite{faghri2018vse++} & 64.61 & \textbf{74.46} & 85.67 & 83.30 & 78.79 & 77.37 \\
		Dual Encoding \cite{dong2019dual} & \textbf{71.92} & 71.52 & 86.12 & 82.00 & 70.59 & 76.43 \\
		Our HGR & 69.48 & 71.21 & \textbf{86.27} & \textbf{84.05} & \textbf{82.04} & \textbf{78.61} \\ \bottomrule
	\end{tabular}
	\vspace{-1em}
\end{table*}

\begin{figure*}
	\includegraphics[width=1\linewidth]{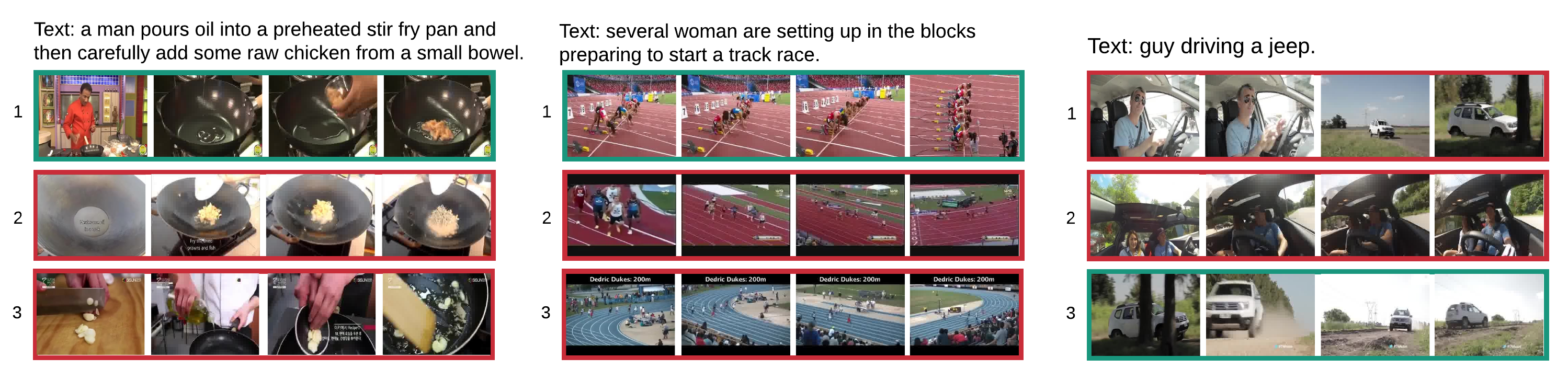}
	\caption{ Text-to-video retrieval examples on MSR-VTT testing set. We visualize top 3 retrieved videos (green: correct; red: incorrect).}
	\label{fig:t2v_examples}
	\vspace{-1.5em}
\end{figure*}

\begin{figure}
	\includegraphics[width=1\linewidth]{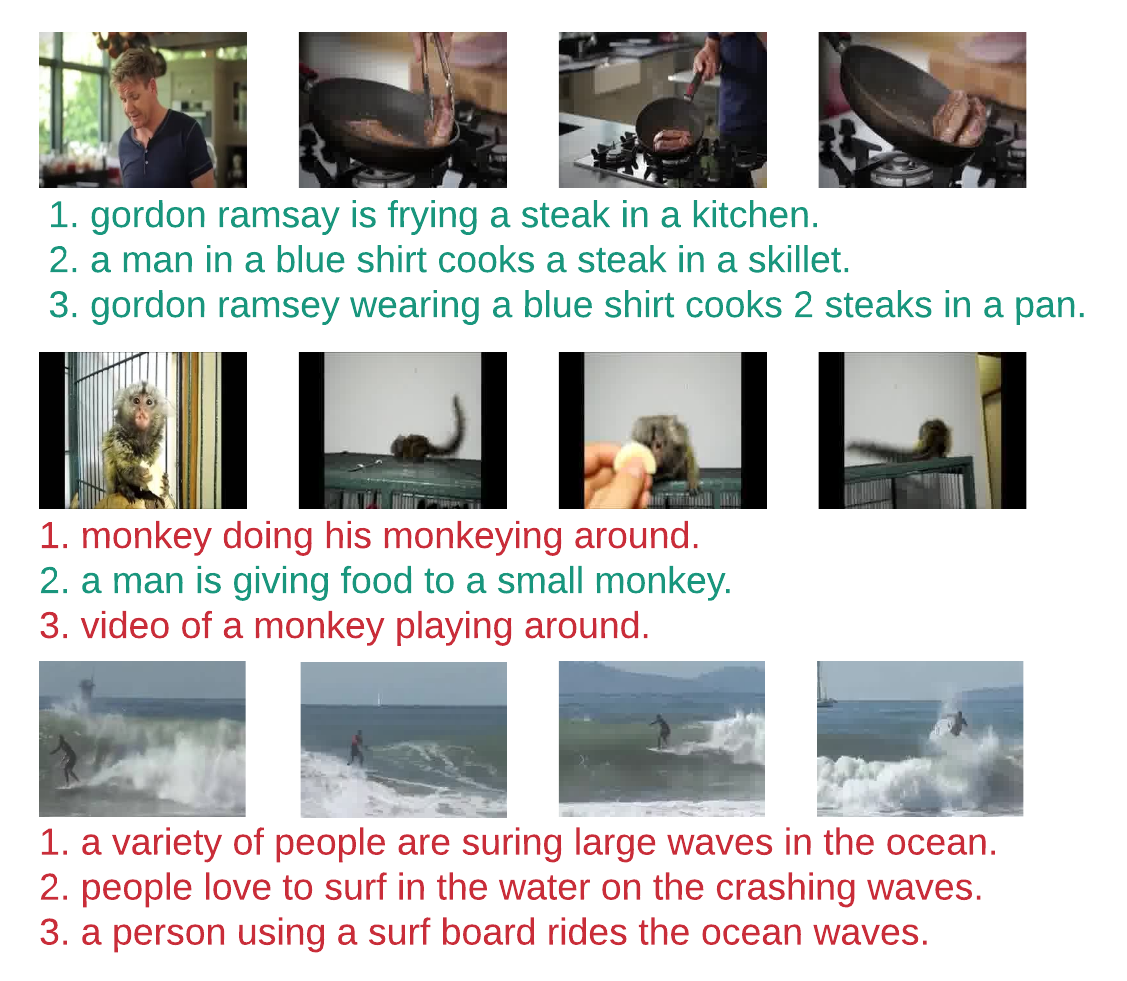}
	\caption{Video-to-text retrieval examples on MSR-VTT testing set with top 3 retrieved texts (green: correct; red: incorrect).}
	\label{fig:v2t_examples}
	\vspace{-1.5em}
\end{figure}

\subsection{Fine-grained Binary Selection}
To prove the ability of our model for fine-grained retrieval, we further propose a binary selection task that requires the model to select a sentence that better matches with a given video from two very similar but semantically different sentences.
We utilize testing videos from the Youtube2Text dataset and randomly select one ground-truth video description for each video as positive sentence.
The negative sentence is generated by perturbing the ground-truth sentence in one of the following ways: 
\parskip=0.1em
\begin{enumerate}[itemsep=0.1em,parsep=0em,topsep=0em,partopsep=0em]
	\item switch roles: switching agent and patient of an action;
	\item replace actions: replacing action with random action;
	\item replace persons: replacing agent or patient entities with random agents or patients;
	\item replace scenes: randomly replacing scene entities;
	\item incomplete events: only keeping part of all actions, entities in the sentence;
\end{enumerate}
We then ask human workers to ensure the automatic generated sentences are syntactically correct but indeed semantically inconsistent with the video content.
Examples can be found in the supplementary material.

Table~\ref{tab:binary_task} presents results in different binary selection tasks.
For the switching roles task, our model outperforms VSE++ model with absolute 4.87\%, but is slightly inferior to Dual Encoding model.
We suspect the reason is that video descriptions in Youtube2Text are relatively short (7 words on average per sentence), which makes sequential models with local contexts such as LSTM, CNN in Dual Encoding model sufficient to capture the event structure.
For the replacing tasks, the HGR model achieves the best performance to distinguish entity replacement especially for scenes.
The largest improvements of our HGR model lies in the incomplete events task, where both the two sentences are relevant to video contents but one captures more details.
Due to the fusion of hierarchical levels from global to local, our model can select the more comprehensive sentence.

\subsection{Qualitative Results}
We visualize some examples on the MSR-VTT testing split for text-to-video retrieval in Figure~\ref{fig:t2v_examples}.
In the left example, our model successfully retrieves the correct video which contains all actions and entities described in the sentence, while the second video only lacks ``pour oil'' action and the third video does not contain ``chicken'' entity. 
In the middle example, the HGR model also distinguishes different relationship of actions such as ``prepare to start a track race'' and ``run in a track race''.
The right example shows a fail case, where the top retrieved videos are largely relevant to the text query though are not ground-truth.
In Figure~\ref{fig:v2t_examples}, we provide qualitative results on video-to-text retrieval as well, which demonstrate the effectiveness of our HGR model for cross-modal retrieval on both directions.

\section{Conclusion}
Most successful cross-modal video-text retrieval systems are based on joint embedding approaches. However, simple embeddings are insufficient to capture fine-grained semantics in complicated videos and texts. 
Therefore, in this work, we propose a Hierarchical Graph Reasoning (HGR) model which decomposes videos and texts into hierarchical semantic levels including events, actions, and entities. 
It then generates hierarchical textual embeddings via attention-based graph reasoning and align texts with videos at different levels.
The overall cross-modal matching is generated by aggregating matching from different levels. 
Superior experimental results on three video-text datasets demonstrate the advantages of our model.
The proposed HGR model also achieves better generalization performance on unseen dataset and is capable of distinguishing fine-grained semantic differences. 


{\small
\bibliographystyle{ieee_fullname}
\bibliography{reference}

\begin{thebibliography}{10}\itemsep=-1pt

\bibitem{i3d}
Joao Carreira and Andrew Zisserman.
\newblock Quo vadis, action recognition? a new model and the kinetics dataset.
\newblock In {\em Proceedings of the IEEE Conference on Computer Vision and
  Pattern Recognition}, pages 6299--6308, 2017.

\bibitem{chang2015semantic}
Xiaojun Chang, Yi Yang, Alexander Hauptmann, Eric~P Xing, and Yao-Liang Yu.
\newblock Semantic concept discovery for large-scale zero-shot event detection.
\newblock In {\em Proceedings of the Twenty-fourth International Joint
  Conference on Artificial Intelligence}, 2015.

\bibitem{dalton2013zero}
Jeffrey Dalton, James Allan, and Pranav Mirajkar.
\newblock Zero-shot video retrieval using content and concepts.
\newblock In {\em Proceedings of the 22nd ACM international conference on
  Information \& Knowledge Management}, pages 1857--1860. ACM, 2013.

\bibitem{dong2018predicting}
Jianfeng Dong, Xirong Li, and Cees~GM Snoek.
\newblock Predicting visual features from text for image and video caption
  retrieval.
\newblock {\em IEEE Transactions on Multimedia}, 20(12):3377--3388, 2018.

\bibitem{dong2019dual}
Jianfeng Dong, Xirong Li, Chaoxi Xu, Shouling Ji, Yuan He, Gang Yang, and Xun
  Wang.
\newblock Dual encoding for zero-example video retrieval.
\newblock In {\em Proceedings of the IEEE Conference on Computer Vision and
  Pattern Recognition}, pages 9346--9355, 2019.

\bibitem{faghri2018vse++}
Fartash Faghri, David~J Fleet, Jamie~Ryan Kiros, and Sanja Fidler.
\newblock Vse++: Improving visual-semantic embeddings with hard negatives.
\newblock In {\em Proceedings of the British Machine Vision Conference}, 2018.

\bibitem{feng2014cross}
Fangxiang Feng, Xiaojie Wang, and Ruifan Li.
\newblock Cross-modal retrieval with correspondence autoencoder.
\newblock In {\em Proceedings of the 22nd ACM international conference on
  Multimedia}, pages 7--16. ACM, 2014.

\bibitem{frome2013devise}
Andrea Frome, Greg~S Corrado, Jon Shlens, Samy Bengio, Jeff Dean, Tomas
  Mikolov, et~al.
\newblock Devise: A deep visual-semantic embedding model.
\newblock In {\em Proceedings of the Advances in Neural Information Processing
  Systems}, pages 2121--2129, 2013.

\bibitem{gu2018look}
Jiuxiang Gu, Jianfei Cai, Shafiq~R Joty, Li Niu, and Gang Wang.
\newblock Look, imagine and match: Improving textual-visual cross-modal
  retrieval with generative models.
\newblock In {\em Proceedings of the IEEE Conference on Computer Vision and
  Pattern Recognition}, pages 7181--7189, 2018.

\bibitem{guadarrama2013youtube2text}
Sergio Guadarrama, Niveda Krishnamoorthy, Girish Malkarnenkar, Subhashini
  Venugopalan, Raymond Mooney, Trevor Darrell, and Kate Saenko.
\newblock Youtube2text: Recognizing and describing arbitrary activities using
  semantic hierarchies and zero-shot recognition.
\newblock In {\em Proceedings of the IEEE International Conference on Computer
  Vision}, pages 2712--2719, 2013.

\bibitem{habibian2014composite}
Amirhossein Habibian, Thomas Mensink, and Cees~GM Snoek.
\newblock Composite concept discovery for zero-shot video event detection.
\newblock In {\em Proceedings of International Conference on Multimedia
  Retrieval}, page~17. ACM, 2014.

\bibitem{he2016deep}
Kaiming He, Xiangyu Zhang, Shaoqing Ren, and Jian Sun.
\newblock Deep residual learning for image recognition.
\newblock In {\em Proceedings of the IEEE Conference on Computer Vision and
  Pattern Recognition}, pages 770--778, 2016.

\bibitem{hochreiter1997long}
Sepp Hochreiter and J{\"u}rgen Schmidhuber.
\newblock Long short-term memory.
\newblock {\em Neural computation}, 9(8):1735--1780, 1997.

\bibitem{hodosh2016focused}
Micah Hodosh and Julia Hockenmaier.
\newblock Focused evaluation for image description with binary forced-choice
  tasks.
\newblock In {\em Proceedings of the 5th Workshop on Vision and Language},
  pages 19--28, 2016.

\bibitem{hu2019evaluating}
Hexiang Hu, Ishan Misra, and Laurens van~der Maaten.
\newblock Evaluating text-to-image matching using binary image selection
  (bison).
\newblock In {\em Proceedings of the IEEE International Conference on Computer
  Vision Workshops}, 2019.

\bibitem{hu2019language}
Ronghang Hu, Anna Rohrbach, Trevor Darrell, and Kate Saenko.
\newblock Language-conditioned graph networks for relational reasoning.
\newblock {\em arXiv preprint arXiv:1905.04405}, 2019.

\bibitem{huang2018learning}
Yan Huang, Qi Wu, Chunfeng Song, and Liang Wang.
\newblock Learning semantic concepts and order for image and sentence matching.
\newblock In {\em Proceedings of the IEEE Conference on Computer Vision and
  Pattern Recognition}, pages 6163--6171, 2018.

\bibitem{keselj2009speech}
Daniel Jurafsky and James~H. Martin.
\newblock Speech and language processing, 2009.

\bibitem{karpathy2015deep}
Andrej Karpathy and Li Fei-Fei.
\newblock Deep visual-semantic alignments for generating image descriptions.
\newblock In {\em Proceedings of the IEEE Conference on Computer Vision and
  Pattern Recognition}, pages 3128--3137, 2015.

\bibitem{kipf2016semi}
Thomas~N Kipf and Max Welling.
\newblock Semi-supervised classification with graph convolutional networks.
\newblock {\em arXiv preprint arXiv:1609.02907}, 2016.

\bibitem{kiros2014unifying}
Ryan Kiros, Ruslan Salakhutdinov, and Richard~S Zemel.
\newblock Unifying visual-semantic embeddings with multimodal neural language
  models.
\newblock {\em arXiv preprint arXiv:1411.2539}, 2014.

\bibitem{lee2018stacked}
Kuang-Huei Lee, Xi Chen, Gang Hua, Houdong Hu, and Xiaodong He.
\newblock Stacked cross attention for image-text matching.
\newblock In {\em Proceedings of the European Conference on Computer Vision},
  pages 201--216, 2018.

\bibitem{li2019visual}
Kunpeng Li, Yulun Zhang, Kai Li, Yuanyuan Li, and Yun Fu.
\newblock Visual semantic reasoning for image-text matching.
\newblock In {\em Proceedings of the IEEE International Conference on Computer
  Vision}, pages 4654--4662, 2019.

\bibitem{li2019relation}
Linjie Li, Zhe Gan, Yu Cheng, and Jingjing Liu.
\newblock Relation-aware graph attention network for visual question answering.
\newblock {\em arXiv preprint arXiv:1903.12314}, 2019.

\bibitem{li2016tgif}
Yuncheng Li, Yale Song, Liangliang Cao, Joel Tetreault, Larry Goldberg,
  Alejandro Jaimes, and Jiebo Luo.
\newblock Tgif: A new dataset and benchmark on animated gif description.
\newblock In {\em Proceedings of the IEEE Conference on Computer Vision and
  Pattern Recognition}, pages 4641--4650, 2016.

\bibitem{Liu19a}
Yang Liu, Samuel Albanie, Arsha Nagrani, and Andrew Zisserman.
\newblock Use what you have: Video retrieval using representations from
  collaborative experts.
\newblock In {\em Proceedings of the British Machine Vision Conference}, 2019.

\bibitem{mithun2018learning}
Niluthpol~Chowdhury Mithun, Juncheng Li, Florian Metze, and Amit~K
  Roy-Chowdhury.
\newblock Learning joint embedding with multimodal cues for cross-modal
  video-text retrieval.
\newblock In {\em Proceedings of the 2018 ACM on International Conference on
  Multimedia Retrieval}, pages 19--27. ACM, 2018.

\bibitem{pennington2014glove}
Jeffrey Pennington, Richard Socher, and Christopher Manning.
\newblock Glove: Global vectors for word representation.
\newblock In {\em Proceedings of the 2014 conference on empirical methods in
  natural language processing}, pages 1532--1543, 2014.

\bibitem{schlichtkrull2018modeling}
Michael Schlichtkrull, Thomas~N Kipf, Peter Bloem, Rianne Van Den~Berg, Ivan
  Titov, and Max Welling.
\newblock Modeling relational data with graph convolutional networks.
\newblock In {\em European Semantic Web Conference}, pages 593--607. Springer,
  2018.

\bibitem{shi2019simple}
Peng Shi and Jimmy Lin.
\newblock Simple bert models for relation extraction and semantic role
  labeling.
\newblock {\em arXiv preprint arXiv:1904.05255}, 2019.

\bibitem{song2019polysemous}
Yale Song and Mohammad Soleymani.
\newblock Polysemous visual-semantic embedding for cross-modal retrieval.
\newblock In {\em Proceedings of the IEEE Conference on Computer Vision and
  Pattern Recognition}, pages 1979--1988, 2019.

\bibitem{sun2018actor}
Chen Sun, Abhinav Shrivastava, Carl Vondrick, Kevin Murphy, Rahul Sukthankar,
  and Cordelia Schmid.
\newblock Actor-centric relation network.
\newblock In {\em Proceedings of the European Conference on Computer Vision},
  pages 318--334, 2018.

\bibitem{velivckovic2017graph}
Petar Veli{\v{c}}kovi{\'c}, Guillem Cucurull, Arantxa Casanova, Adriana Romero,
  Pietro Lio, and Yoshua Bengio.
\newblock Graph attention networks.
\newblock {\em arXiv preprint arXiv:1710.10903}, 2017.

\bibitem{vendrov2015order}
Ivan Vendrov, Ryan Kiros, Sanja Fidler, and Raquel Urtasun.
\newblock Order-embeddings of images and language.
\newblock {\em arXiv preprint arXiv:1511.06361}, 2015.

\bibitem{wang2019neighbourhood}
Peng Wang, Qi Wu, Jiewei Cao, Chunhua Shen, Lianli Gao, and Anton van~den
  Hengel.
\newblock Neighbourhood watch: Referring expression comprehension via
  language-guided graph attention networks.
\newblock In {\em Proceedings of the IEEE Conference on Computer Vision and
  Pattern Recognition}, pages 1960--1968, 2019.

\bibitem{wang2018videos}
Xiaolong Wang and Abhinav Gupta.
\newblock Videos as space-time region graphs.
\newblock In {\em Proceedings of the European Conference on Computer Vision},
  pages 399--417, 2018.

\bibitem{Wang_2019_ICCV}
Xin Wang, Jiawei Wu, Junkun Chen, Lei Li, Yuan-Fang Wang, and William~Yang
  Wang.
\newblock Vatex: A large-scale, high-quality multilingual dataset for
  video-and-language research.
\newblock In {\em Proceedings of the IEEE International Conference on Computer
  Vision}, October 2019.

\bibitem{wang2019vatex}
Xin Wang, Jiawei Wu, Junkun Chen, Lei Li, Yuan-Fang Wang, and William~Yang
  Wang.
\newblock Vatex: A large-scale, high-quality multilingual dataset for
  video-and-language research.
\newblock {\em Proceedings of the IEEE International Conference on Computer
  Vision}, 2019.

\bibitem{Wray_2019_ICCV}
Michael Wray, Diane Larlus, Gabriela Csurka, and Dima Damen.
\newblock Fine-grained action retrieval through multiple parts-of-speech
  embeddings.
\newblock In {\em Proceedings of the IEEE International Conference on Computer
  Vision}, October 2019.

\bibitem{wu2019unified}
Hao Wu, Jiayuan Mao, Yufeng Zhang, Yuning Jiang, Lei Li, Weiwei Sun, and
  Wei-Ying Ma.
\newblock Unified visual-semantic embeddings: Bridging vision and language with
  structured meaning representations.
\newblock In {\em Proceedings of the IEEE Conference on Computer Vision and
  Pattern Recognition}, pages 6609--6618, 2019.

\bibitem{xu2016msr}
Jun Xu, Tao Mei, Ting Yao, and Yong Rui.
\newblock Msr-vtt: A large video description dataset for bridging video and
  language.
\newblock In {\em Proceedings of the IEEE conference on Computer Vision and
  Pattern Recognition}, pages 5288--5296, 2016.

\bibitem{yang2018graph}
Jianwei Yang, Jiasen Lu, Stefan Lee, Dhruv Batra, and Devi Parikh.
\newblock Graph r-cnn for scene graph generation.
\newblock In {\em Proceedings of the European Conference on Computer Vision},
  pages 670--685, 2018.

\bibitem{yu2018joint}
Youngjae Yu, Jongseok Kim, and Gunhee Kim.
\newblock A joint sequence fusion model for video question answering and
  retrieval.
\newblock In {\em Proceedings of the European Conference on Computer Vision},
  pages 471--487, 2018.

\end{thebibliography}
}

\appendix
\section{Semantic Roles}
We parse text into hierarchical semantic role graph consisting of global event node, action nodes and entity nodes.
The action nodes are connected with global event node with edge type of action, and the entity nodes are linked with the corresponding action nodes with different edge types according to their semantic roles.
Table~\ref{tab:semantic_roles} presents all semantic roles used in our graph and their descriptions based on linguistic experts \cite{keselj2009speech}.

\begin{table}[hbpt]
	\centering
	\caption{Semantic roles in parsed semantic role graph.}
	\label{tab:semantic_roles}
	\begin{tabular}{l|l} \toprule
		Semantic Role & Description \\ \midrule
		Event & global event description   \\ \midrule
		Action & verb \\ \midrule
		ARG0 & proto-agent \\
		ARG1 & proto-patient  \\
		ARG2 & instrument, benefactive  \\
		ARG3 & start point   \\
		ARG4 & end point  \\
		ARGM-LOC & location (where) \\
		ARGM-MNR & manner (how)  \\
		ARGM-TMP & time (when)  \\
		ARGM-DIR & direction (where to/from)  \\
		ARGM-ADV & miscellaneous   \\
		OTHERS & other argument types   \\ \bottomrule
	\end{tabular}
\end{table}

\begin{figure}[hbpt]
    \centering
    \includegraphics[width=1\linewidth]{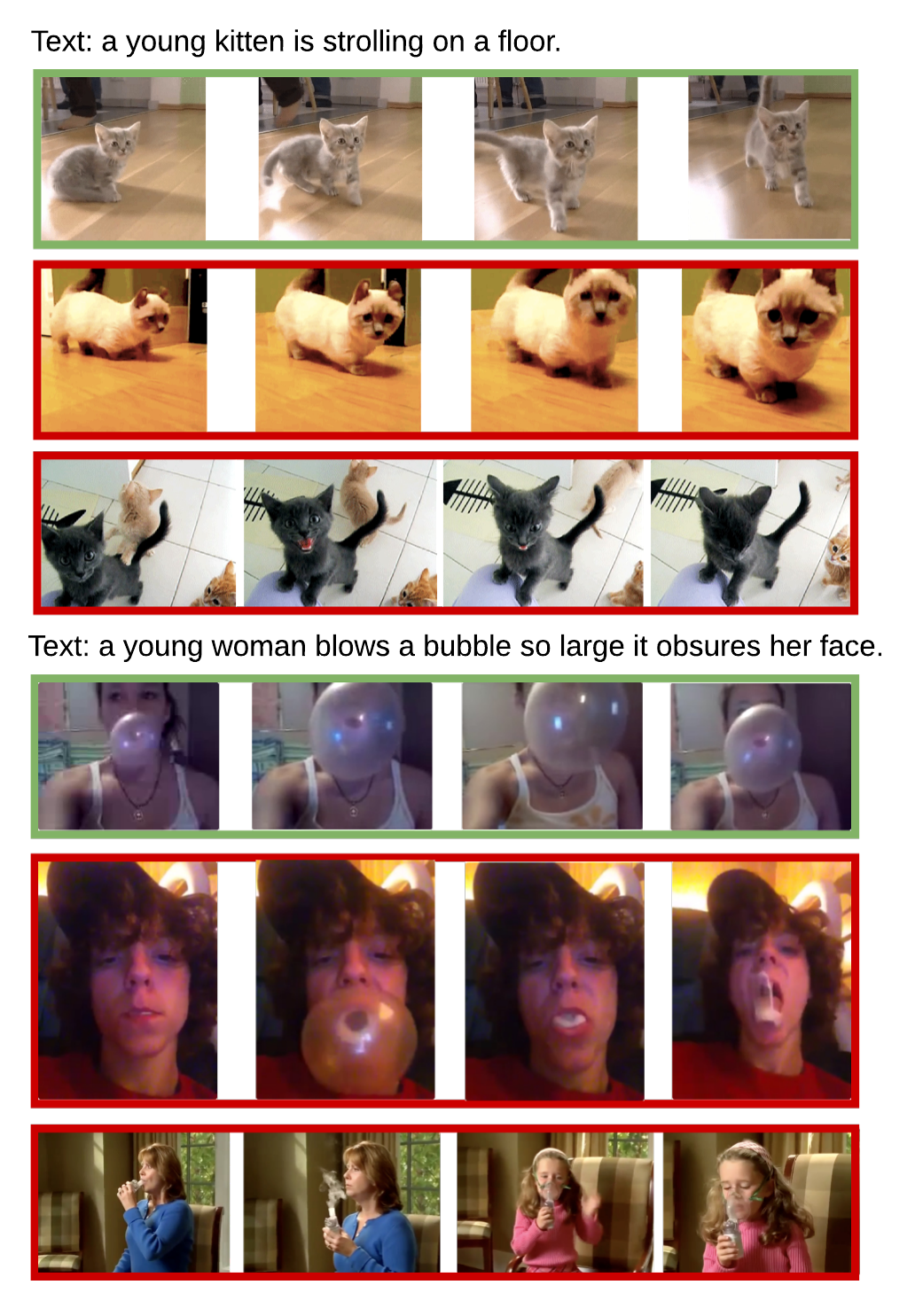}
    \caption{Cross-modal video-text retrieval results on TGIF and VATEX testing set.}
    \label{fig:supp_t2v_examples}
\end{figure}

\section{Binary Selection Task}
In order to evaluate fine-grained discrimination ability on texts of different video-text retrieval models, we propose a binary selection task which requires the model to select a sentence that better matches with a given video from two very similar but semantically different sentences. 
We utilize testing videos from the Youtube2Text dataset and randomly select one ground-truth video description for each video as the positive sentence. The negative sentence is generated by perturbing the ground-truth sentence as shown in Table~\ref{tab:binary_selection_task}. 

\begin{table*}[h]
\centering
\caption{Examples of different perturbation types in binary selection task.}
\label{tab:binary_selection_task}
\begin{tabular}{l|l|l} \toprule
Task & Description & Example \\  \midrule
\multirow{2}{*}{\begin{tabular}[c]{@{}l@{}}switch \\ roles\end{tabular}} & \multirow{2}{*}{\begin{tabular}[c]{@{}l@{}}switching the agent and patient \\ of an action in the sentence\end{tabular}} & positive: a woman is cutting an onion.\\
 &  &  negative: an onion is cutting a woman.\\ \midrule
\multirow{2}{*}{\begin{tabular}[c]{@{}l@{}}replace \\ actions\end{tabular}} & \multirow{2}{*}{\begin{tabular}[c]{@{}l@{}}replacing action in the sentence\\  with a random one\end{tabular}} & positive: a person pours coconut water into a bowl.\\
 &  &  negative:  a person drives coconut water into a bowl. \\ \midrule
\multirow{2}{*}{\begin{tabular}[c]{@{}l@{}}replace \\ persons\end{tabular}} & \multirow{2}{*}{\begin{tabular}[c]{@{}l@{}}replacing agent or patient in the\\  sentence with a random one\end{tabular}} & positive: a man is keep the knife on the machine. \\
 &  &  negative: a man is keep a dog on the floor on the machine. \\ \midrule
\multirow{2}{*}{\begin{tabular}[c]{@{}l@{}}replace \\ scenes\end{tabular}} & \multirow{2}{*}{\begin{tabular}[c]{@{}l@{}}replacing scene in the sentence\\  with a random one\end{tabular}} & positive: a man strums a violin on a stage. \\
 &  &  negative: a man strums a violin in the beach. \\ \midrule
\multirow{2}{*}{\begin{tabular}[c]{@{}l@{}}incomplete \\ events\end{tabular}} & \multirow{2}{*}{\begin{tabular}[c]{@{}l@{}}only keeping part of the description \\ of an event \end{tabular}} & positive: men are dancing in towels.  \\
 &  &  negative: men in towels. \\ \bottomrule
\end{tabular}
\end{table*}

\section{Additional Qualitative Examples}
We visualize some examples on cross-modal video-text retrieval on TGIF and VATEX datasets in Figure~\ref{fig:supp_t2v_examples}.
Our model achieves robust and superior performance on different datasets for cross-modal retrieval.

\end{document}